\documentclass{article}

\PassOptionsToPackage{numbers, compress}{natbib}

\usepackage[preprint]{neurips_2025}

\usepackage{wrapfig}

\usepackage[utf8]{inputenc} 
\usepackage[T1]{fontenc}    
\usepackage{hyperref}       
\usepackage{url}            
\usepackage{booktabs}       
\usepackage{amsfonts}       
\usepackage{nicefrac}       
\usepackage{microtype}      
\usepackage{xcolor}         
\usepackage{amsmath} 
\usepackage{multirow}
\usepackage{graphicx}
\usepackage{float}
\usepackage{natbib}

\title{EGFormer: Towards Efficient and Generalizable Multimodal Semantic Segmentation}

\author{
Zelin Zhang$^{1}$ \quad Tao Zhang$^{2}$ \quad KediLI$^{1}$ \quad
Xu Zheng$^{3}$\thanks{Corresponding Author: Xu Zheng <\texttt{zhengxu128@gmail.com}>} \\
\\
$^{1}$University of Sydney\quad 
$^{2}$University of Technology Sydney \quad
$^{3}$HKUST(GZ)
}

\begin{document}
\maketitle
\begin{abstract}
Research endeavors have been made on multimodal semantic segmentation using various backbone architectures. However, while most methods primarily focus on improving performance, their computational efficiency remains insufficiently explored.
To bridge this gap, we introduce an efficient multimodal semantic segmentation framework named EGFormer, which flexibly integrates an arbitrary number of modalities while significantly reducing model parameters and inference time without sacrificing performance. Specifically, our framework consists of two novel modules. First, an Any-modal Scoring Module (ASM) assigns importance scores to each modality independently, enabling dynamic ranking based on their feature maps. On top, a Modal Dropping Module (MDM) dynamically filters out less informative modalities at each stage, selectively preserving and aggregating only the most valuable information. This strategy enables the model to effectively utilize useful features from all available modalities while discarding redundant ones, ensuring high segmentation quality. 
Beyond computational efficiency, we further test the synthetic-to-real task with our EGFormer to show the generalizability. 
Extensive experiments demonstrate that our model achieves competitive performance with up to \textbf{\textit{88\% reduction in parameters and 50\% fewer GFLOPs}}. Meanwhile,
under unsupervised domain adaptation settings, our EGFormer achieves state-of-the-art transfer performance across SoTA methods.
\end{abstract}

\section{Introduction}
\label{sec:introduction}
With the rapid development of sensor platforms (e.g., autonomous vehicles, drones, and augmented reality devices), multimodal semantic segmentation has garnered increasing attention as a critical means to enhance scene understanding \cite{alonso2019evsegnet,cao2021shapeconv,chen2020bidirectional,jia2023event,sun2019rtfnet,zhang2023delivering,lyu2024omnibind}. This task integrates data from diverse sensors (e.g., LiDAR point clouds and event streams), significantly improving the robustness and accuracy of semantic segmentation in complex environments.

Early research predominantly focused on constructing specialized fusion architectures for fixed modality combinations, such as RGB-Depth \cite{sun2019rtfnet,hazirbas2016fusenet}, RGB-LiDAR \cite{elmadawi2019rgb,li2023mseg3d,lyu2024unibind}, and RGB-Event \cite{yao2024sam,zhao2025eseg,zheng2024eventdance,zhou2024exact}. While these methods achieved strong performance under predefined modality settings, they lack scalability and struggle to adapt to dynamically changing modality combinations in real-world scenarios. Consequently, studies on arbitrary-modal fusion have emerged \cite{zhang2023delivering,brodermann2025cafuser,li2024stitchfusion,zheng2024centering}, aiming to develop generalizable frameworks for flexible sensor combinations. Among these, CMNeXt \cite{zhang2023delivering} introduced a dual-branch heterogeneous architecture: a primary branch processes the RGB modality, while an auxiliary branch integrates additional modalities (e.g., depth, event, LiDAR). Its core module, Self-Query Hub, dynamically selects informative features from auxiliary modalities for fusion with the primary branch, achieving robust performance across diverse modality combinations. Subsequent efforts \cite{brodermann2025cafuser,zheng2024centering,wang2022multimodal} have explored single-branch shared architectures or dynamic pathway mechanisms to unify arbitrary-modal inputs, aiming to enhance cross-modal interaction efficiency. For instance, MAGIC \cite{zheng2024centering} stacks all modality features and processes them through attention mechanisms, while others leverage Transformer frameworks for modality-agnostic representation learning. Although these strategies improve accuracy, they often suffer from structural complexity and high computational costs, particularly exhibiting significant inference latency.

While arbitrary-modal fusion methods advance segmentation accuracy, their architectural designs prioritize performance at the expense of lightweight deployment efficiency. This creates a critical bottleneck for real-world applications in resource-constrained scenarios (e.g., edge devices or real-time systems), where existing methods—reliant on heavy network structures—impose prohibitive computational overhead. The inherent trade-off between "high performance" and "high cost" remains a pivotal challenge hindering the practical adoption of arbitrary-modal semantic segmentation.

\begin{figure}
  \centering
  \includegraphics[width=0.85\linewidth]{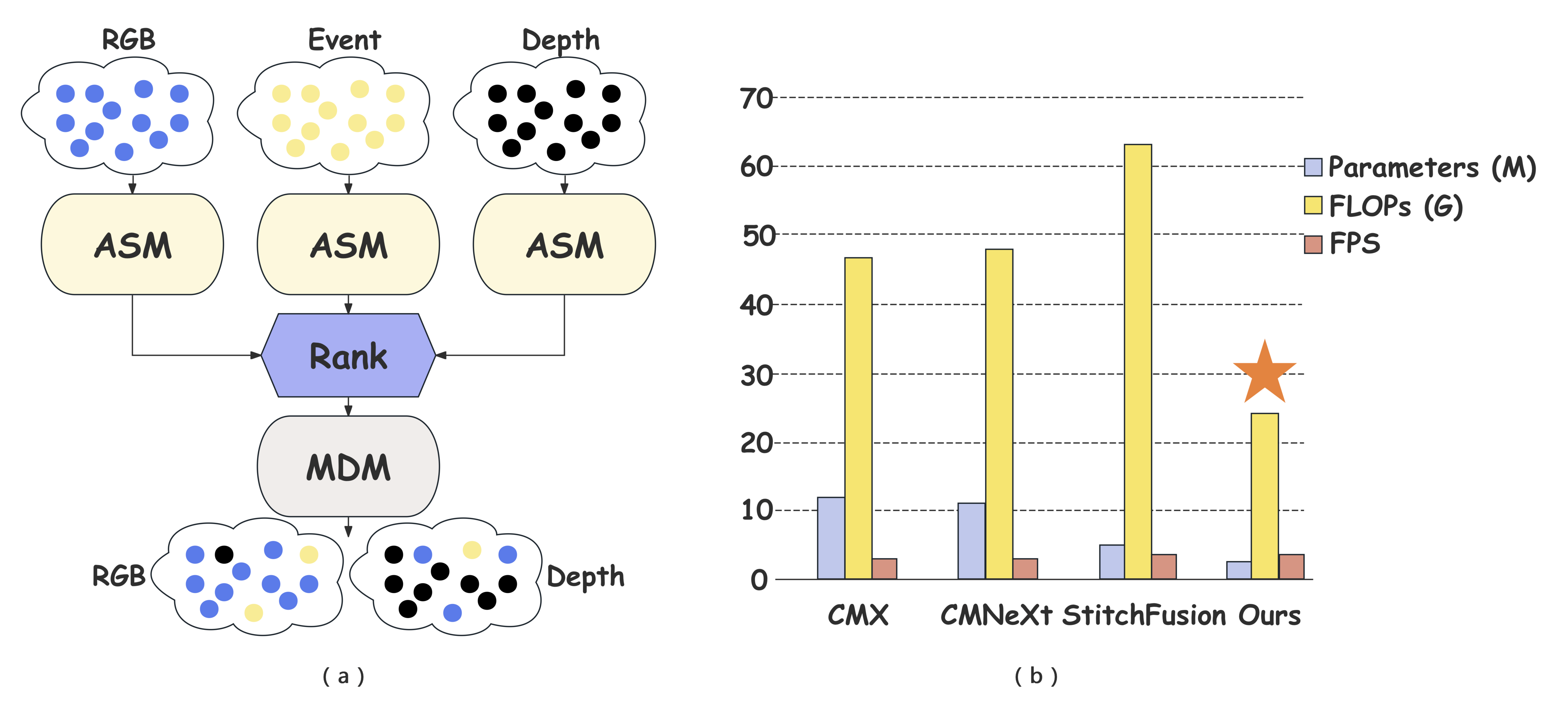}
  \caption{(a) Multi-modal data are processed by EGFormer, where robust modality features are selected and fine-tuned. (b) Model efficiency comparison across different methods on DELVIER~\cite{zhang2023delivering}.}
  \vspace{-20pt}
  \label{cover figure}
\end{figure}

To tackle these challenges, we propose EGFormer, a lightweight and modular framework for arbitrary-modal semantic segmentation as illustrated in Figure~\ref{cover figure} (a). Our design is both efficient and flexible, capable of adapting to varying modality combinations with minimal computational overhead. At the core of EGFormer are two novel modules. The Any-modal Scoring Module (ASM) dynamically evaluates and ranks each modality based on its task relevance, enabling the model to prioritize informative signals. On top of this, the Modality Dropping Module (MDM) selectively suppresses redundant or noisy modalities—not by discarding them, but by retaining and redistributing their useful information. This dynamic fusion strategy ensures robust performance under partial sensor degradation and avoids the cost of processing all modalities equally.

Beyond computational efficiency, our method also demonstrates strong generalization under synthetic-to-real domain adaptation~\cite{zheng2024360sfuda++,zheng2024semantics,zheng2023both,zheng2023look}. We integrate a multimodal consistency voting strategy into a teacher-student framework to generate reliable pseudo-labels from diverse modality subsets. This mechanism supports learning from unlabeled real-world data without requiring full modality availability, making our approach especially practical for field deployment.

We extensively evaluate EGFormer on the MCubeS \cite{liang2022multimodal}, MUSES \cite{muses} and DELIVER \cite{zhang2023delivering} benchmarks, covering 15 modality combinations from single-modal to quad-modal settings. Results show that our model consistently outperforms the strong baseline CMNeXt in 11 out of 15 configurations, with up to 88\% fewer parameters and 50\% lower GFLOPs, while maintaining or exceeding segmentation accuracy, see Figure~\ref{cover figure} (b).
Our main contributions are: \textbf{(I)} We propose an efficient and flexible framework with dynamic modality scoring and redundancy-aware suppression, enabling robust segmentation under missing or degraded sensors. \textbf{(II)} We introduce a pseudo-label voting strategy within a teacher-student framework, allowing cross-domain adaptation without requiring modality completeness. \textbf{(III)} We achieve strong performance on DELIVER and MCubeS under varying modality settings, outperforming state-of-the-art methods with significantly reduced complexity.

\section{Related work}
\label{sec:related work}
\paragraph{Semantic Segmentation.} 
As one of the fundamental tasks in computer vision, semantic segmentation has demonstrated extensive application potential in multiple fields such as remote sensing imagery \cite{audebert2018beyond,dimitrovski2024unet,wang2024metasegnet,zheng2024transformer,DBLP:journals/pr/ZhengLZW25,DBLP:journals/pr/ChenDZZM24} and robot perception \cite{munasinghe2022covered}. Early semantic segmentation methods were mostly based on convolutional neural networks. For example, FCN \cite{long2015fully} first proposed using a fully convolutional network for end-to-end pixel-level prediction. Subsequent research has further advanced performance by focusing on multi-scale feature extraction \cite{chen2017deeplab,chen2018encoder,hou2020strip,zhao2017pyramid}, the integration of attention mechanisms \cite{choi2020cars,fu2019dual,huang2019ccnet,yuan2021ocnet}, enhancing the delineation of object boundaries \cite{borse2021inverseform,ding2019boundary,li2020improving}. Recently, Transformer-based methods have emerged as strong alternatives for semantic segmentation, demonstrating excellent capabilities in modeling global dependencies and delivering strong performance across various benchmarks. Early works such as DPT \cite{ranftl2021vision} leveraged Vision Transformers to replace traditional convolutional backbones for dense prediction tasks. Following this, subsequent research \cite{gu2022multi,guo2022segnext,liu2022swin,liu2021swin,strudel2021segmenter,xie2021segformer,zhang2022segvit,zheng2021rethinking} has further advanced Transformer-based segmentation models through hierarchical encoding, sequence modeling, and hybrid architectures combining convolutions and attention mechanisms. However, despite their remarkable accuracy, these approaches often suffer from heavy computational overhead and limited efficiency, posing challenges for deployment in resource-constrained environments. In this work, we focus on improving computational efficiency while maintaining competitive segmentation performance.
\paragraph{Multimodal Semantic Segmentation.} 
Early multimodal semantic segmentation methods \cite{cao2021shapeconv,chen2020bidirectional,sun2019rtfnet,hazirbas2016fusenet,audebert2018beyond,wang2024metasegnet,borse2023xalign,chen2022modality,hui2023bridging,liao2022cross,mei2022glass,pang2023caver,liao2025benchmarking} mainly focused on combining RGB images with a single auxiliary modality to extract complementary features. With the development of new types of sensors, researchers \cite{zhang2023delivering,brodermann2025cafuser,li2024stitchfusion,zheng2024centering,wang2022multimodal,liang2022multimodal,zheng2025reducing,liao2025memorysam,zhao2025unveiling,zheng2024learning,zheng2024magic++,zhu2024customize,zheng2023deep} have begun to explore the expansion from dual-modal to multi-modal fusion, further enhancing the robustness and generalization ability of the system in complex environments. Representative works such as CMNeXt \cite{zhang2023delivering} have achieved flexible semantic segmentation under any modal combination. However, most of these methods default to RGB as the core mode and fail to fully address the problem of performance degradation of the RGB mode under extreme conditions such as low light and bad weather, resulting in obvious vulnerability of the entire system when the sensor fails or the mode is missing. In response to the above challenges, this paper proposes a lightweight and modal-independent multimodal semantic segmentation framework. Through an efficient information fusion mechanism, excellent segmentation performance can be achieved in a multimodal environment with only a very small number of parameters. Furthermore, the proposed method can still maintain high robustness when facing the absence or degradation of modalities, fully verifying its potential in practical application scenarios. Moreover, collecting and annotating large-scale multimodal datasets is costly. To reduce reliance on labeled data, we also evaluate our method under unsupervised domain adaptation settings, where it maintains high performance, demonstrating its practical value.

\section{Methodology}
\label{sec:methodology}
\begin{figure}[t!]
  \centering
  \includegraphics[width=0.85\linewidth, height=7.5cm]{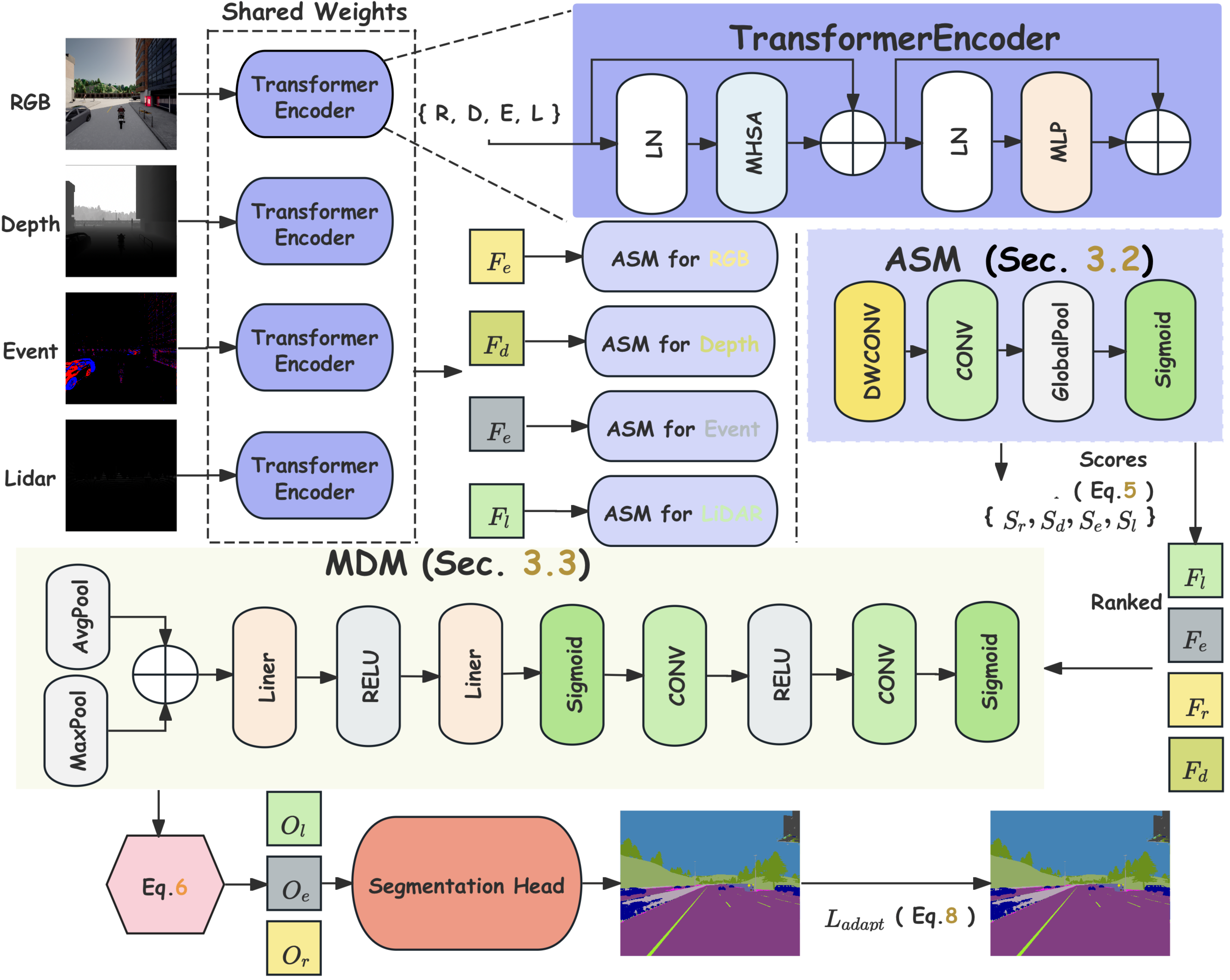}
  \caption{Overall framework of our proposed EGFormer.}
  \vspace{-16pt}
  \label{Architecture}
\end{figure}

In this section, we introduce our efficient and generalizable multimodal semantic segmentation framework, namely, \textbf{\textit{EGFormer}}. As in Figure~\ref{Architecture}, it consists of two modules: the Any-modal Scoring Module (ASM) and the Modality Drop Module (MDM). By effectively exploiting multiple visual modalities, our model achieves a better trade-off between \textbf{\textit{performance}} and \textbf{\textit{computational efficiency}}.

\subsection{Multi-modal input and feature extraction}
\noindent \textbf{Inputs:} EGFormer is designed to flexibly handle an arbitrary number of input modalities. For example, we adopt the four-modal configuration provided by the DELIVER dataset, including RGB images $\mathbf{R} \in \mathbb{R}^{h \times w \times C^R}$, depth maps $\mathbf{D} \in \mathbb{R}^{h \times w \times C^D}$, LiDAR projections $\mathbf{L} \in \mathbb{R}^{h \times w \times C^L}$, and event stacks $\mathbf{E} \in \mathbb{R}^{h \times w \times C^E}$. All modalities have the same spatial resolution of $1024 \times 1024$, and their channel dimensions are unified as $C^R = C^D = C^L = C^E = 3$. The input is structured as a list of modality tensors $\{ \mathbf{R}, \mathbf{D}, \mathbf{L}, \mathbf{E} \}$, which are simultaneously passed into the model.
\textbf{Outputs:} EGFormer adopts a four-stage hierarchical encoder architecture. At each stage, the input modalities are first passed through shared weights transformer encoders to extract deep features, denoted as $\{F_r, F_d, F_e, F_l\}$. These modality-specific features are then passed to ASM, which predicts a confidence score $\{S_r, S_d, S_e, S_l\}$ for each modality to quantify its relative importance. Based on this ranking, MDM identifies and suppresses the least informative modality. Rather than discarding its features entirely, MDM selectively attenuates redundant information while preserving and redistributing its useful components across the remaining modalities. The enhanced outputs $\{O_r, O_d, O_e\}$ are then aggregated and forwarded to the segmentation head for final prediction. This dynamic dropping and redistribution strategy improves robustness under modality degradation, while maintaining efficiency and feature diversity across the network. 

\subsection{Any-modal Scoring Module (ASM)}
To effectively evaluate the importance of each modality in multi-modal inputs, we propose an Any-modal Scoring Module (ASM), see Figure~\ref{Architecture}. This module dynamically estimates the contribution of each modality during the fusion process and identifies the most redundant one, enabling more efficient feature integration. ASM assigns an independent score to each modality feature and aggregates these scores across the batch dimension. The modality with the lowest average score is then regarded as redundant and subsequently suppressed. 
Given $N$ input modalities, their feature maps after encoder processing are denoted as $\{\mathbf{F}_k\}_{k=1}^N$, where $\mathbf{F}_k \in \mathbb{R}^{B \times C \times H \times W}$. Each modality feature is first compressed into a global representation via global average pooling:
\begin{equation}
\setlength{\abovedisplayskip}{3pt}
\setlength{\belowdisplayskip}{3pt}
\mathbf{f}_k = \frac{1}{H \times W} \sum_{h=1}^H \sum_{w=1}^W \mathbf{F}_k^{(h,w)} \in \mathbb{R}^{C}.
\end{equation}

A non-linear projection function $\phi(\cdot)$, implemented as a lightweight convolutional network followed by a sigmoid activation, maps $\mathbf{f}_k$ to a scalar importance score:
\begin{equation}
\setlength{\abovedisplayskip}{3pt}
\setlength{\belowdisplayskip}{3pt}
s_k = \phi(\mathbf{f}_k) = \sigma\left( \mathbf{w}^\top \psi(\mathbf{f}_k) + b \right),
\end{equation}
where $\psi(\cdot)$ denotes a learnable transformation (e.g., depthwise separable convolution), $\mathbf{w} \in \mathbb{R}^m$ and $b \in \mathbb{R}$ are the learned parameters, and $\sigma$ is the sigmoid function ensuring $s_k \in (0, 1)$.
To robustly measure modality significance across the batch, we compute the average score per modality:
\begin{equation}
\setlength{\abovedisplayskip}{3pt}
\setlength{\belowdisplayskip}{3pt}
\bar{s}_k = \frac{1}{B} \sum_{i=1}^{B} s_k^{(i)}, \quad k = 1, \dots, N.
\end{equation}

The least informative modality is then selected by:
\begin{equation}
\setlength{\abovedisplayskip}{3pt}
\setlength{\belowdisplayskip}{3pt}
d = \arg\min_k \bar{s}_k,
\end{equation}
and can be optionally excluded during fusion.

Thus, we formalize the modality scoring process as:
\begin{equation}
d = \arg\min_k \left( \frac{1}{B} \sum_{i=1}^{B} \sigma\left( \mathbf{w}^\top \cdot \psi\left( \frac{1}{HW} \sum_{h,w} \mathbf{F}_k^{(i,h,w)} \right) + b \right) \right),
\end{equation}
which adaptively determines the most redundant modality $d$ in a differentiable and efficient manner.

\begin{wrapfigure}{r}{8cm}
  \centering
  \includegraphics[width=\linewidth]{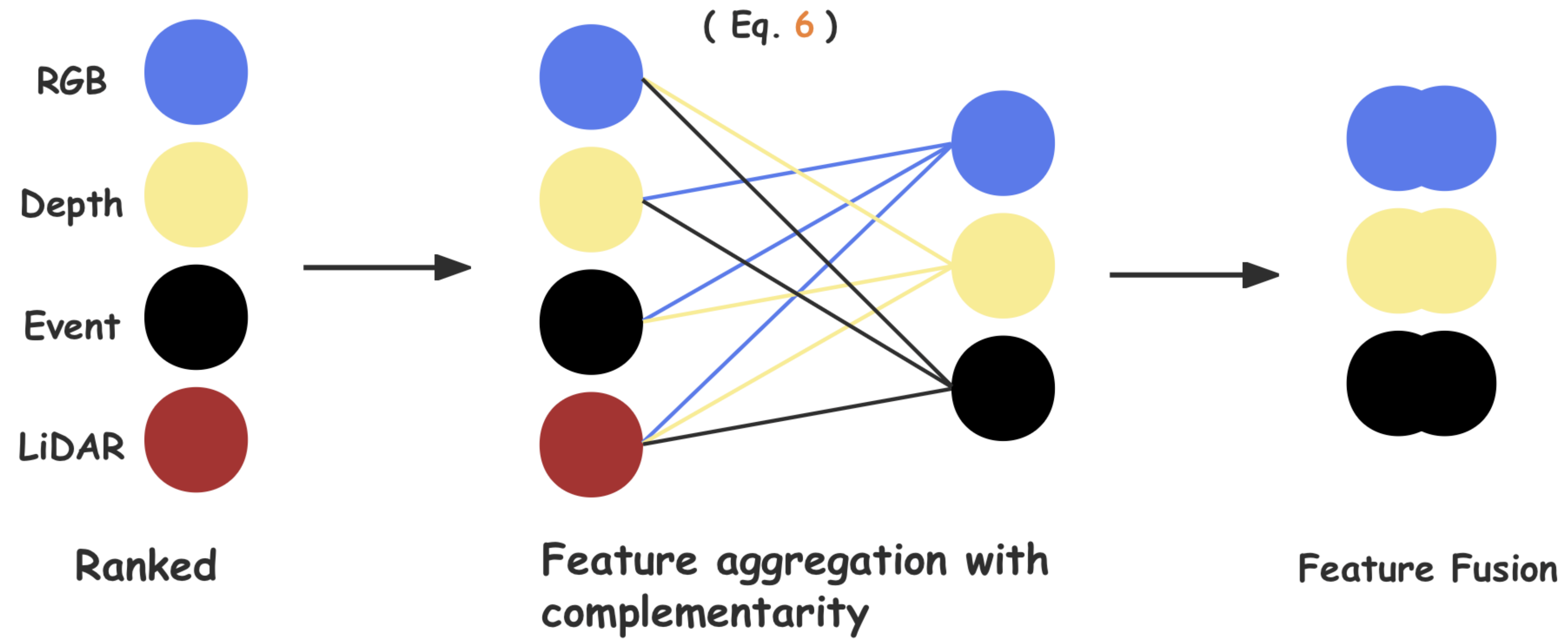}
  \caption{Illustration of MDM module.}
  \label{Illustration of the MDM to drop and compensation information.}
\end{wrapfigure}
\subsection{Modality Dropping Module (MDM)}
Building upon the importance-aware ranking obtained from ASM, we further introduce the Modality Dropping Module (MDM) to enhance model robustness against modality perturbations, as shown in Figure~\ref{Architecture}.
At each stage, MDM identifies the modality with the lowest MSM score, which is then dropped—meaning that its features are excluded from direct fusion but remain involved in feature interaction via cross-modal enhancement.
To effectively compensate for the dropped modality, MDM adopts dual attention strategies: one globally measures the relative significance of different feature dimensions, and the other adaptively emphasizes spatially discriminative regions within the retained modalities.

Formally, given the input modalities ${f_m}$, the remaining modalities after dropping are refined through weighted interactions with the dropped modality. This design allows the network to selectively preserve important signals from the dropped modality while suppressing redundant or noisy information, promoting robust multimodal feature learning under modality degradation. Given the set of modalities $\{f_m\}$, the remaining modalities after dropping are refined by weighted interactions, see Figure~\ref{Illustration of the MDM to drop and compensation information.}:
\begin{equation}
\setlength{\abovedisplayskip}{3pt}
\setlength{\belowdisplayskip}{3pt}
\hat{f}_i = f_i + 0.5 \sum_{j \neq i} W_{c,j} \odot f_j + 0.5 \sum_{j \neq i} W_{s,j} \odot f_j,
\end{equation}
where $W_c$ and $W_s$ are the learned channel and spatial weights and $\odot$ denotes element-wise multiplication.  
By progressively dropping the least reliable modality at each stage, the model is encouraged to rely on complementary information and maintains high performance even when faced with missing or degraded inputs.

\subsection{Synthetic-to-Real domain adaptation with multimodal pseudo-label fusion}
In addition to improving computational efficiency, EGFormer is designed to tackle the challenging synthetic-to-real adaptation setting. To bridge the domain gap between synthetic and real data, we propose a robust pseudo-label generation strategy that leverages multimodal consistency voting. This approach aims to produce reliable pseudo-labels for unlabeled real-world data by aggregating predictions from diverse modality combinations, thereby reducing noise introduced by domain shifts.

Given a set of available modalities $\mathcal{M} = \{ m_r, m_d, m_e, m_l \}$ (e.g., RGB, depth, event, LiDAR), we systematically sample $N$ modality combinations $\{ C_i \}_{i=1}^{N}$, where each $C_i \subseteq \mathcal{M}$ contains a subset of modalities. 
For each combination $C_i$, the teacher model $f_t$ generates a segmentation probability map $P_i \in \mathbb{R}^{B \times C \times H \times W}$, where $C$ denotes the number of classes. To fuse these predictions into a unified pseudo-label, we adopt a \textbf{majority voting scheme}:

\textbf{Class Voting}: For each spatial position $(h, w)$, we compute the argmax class prediction from each $P_i$, yielding $N$ candidate labels.

\textbf{Consistency Filtering}: A position is assigned a pseudo-label only if at least $T$ predictions (e.g., $T = 2$) agree on the class. Formally:
  
\begin{equation}
\setlength{\abovedisplayskip}{3pt}
\setlength{\belowdisplayskip}{3pt}
\hat{y}^{(h,w)} =
\begin{cases}
\text{mode}\left(\left\{ \arg\max P_i^{(h,w)} \right\}_{i=1}^N\right), & \text{if } \text{count}(\text{mode}) \geq T \\
255, & \text{otherwise (ignore label)}
\end{cases}
\end{equation}

Here, 255 marks uncertain regions excluded from training, ensuring that only high-confidence predictions guide the adaptation process.
The voting mechanism aligns with the consensus principle in ensemble learning, where multimodal redundancies suppress modality-specific noise. By requiring agreement across multiple combinations, our method prioritizes class predictions robust to both domain shifts and missing modalities. This contrasts with confidence-based fusion (e.g., weighting by prediction entropy), which may over-rely on biased single-modality confidences. These pseudo-labels serve as supervision for training a student model in a teacher-student framework, as described below.

\subsection{Teacher-Student framework and training objective}
This pseudo-label generation module is embedded in a teacher-student framework, where the teacher’s fused labels supervise the student model on real data. The student is trained using a cross-entropy loss $\mathcal{L}_{\text{adapt}}$ that ignores uncertain regions:
\begin{equation}
\setlength{\abovedisplayskip}{3pt}
\setlength{\belowdisplayskip}{3pt}
\mathcal{L}_{\text{adapt}} = - \frac{1}{|\Omega|} \sum_{(h,w) \in \Omega} \log f_s^{(h,w)}\left( \hat{y}^{(h,w)} \right)
\end{equation}
where $\Omega = \{ (h, w) \mid \hat{y}^{(h,w)} \neq 255 \}$.
To further encourage consistency between teacher and student predictions, we introduce an additional KL divergence loss:
\begin{equation}
\setlength{\abovedisplayskip}{3pt}
\setlength{\belowdisplayskip}{3pt}
\mathcal{L}_{\text{KL}} = \frac{1}{|\Omega|} \sum_{(h,w) \in \Omega} D_{\text{KL}}\left( f_t^{(h,w)} \parallel f_s^{(h,w)} \right)
\label{eq:kl-loss}
\end{equation}

The total training objective becomes:
\begin{equation}
\mathcal{L}_{\text{total}} = \mathcal{L}_{\text{adapt}} + \mathcal{L}_{\text{KL}}
\label{eq:total-loss}
\end{equation}

\section{Experiments}
\label{sec:experiments}
\subsection{Datasets and implementation details}
\textbf{MCubeS}~\cite{liang2022multimodal} is a multimodal material segmentation dataset containing 20 semantic classes. It provides paired images of RGB, Near-Infrared (NIR), Degree of Linear Polarization (DoLP), and Angle of Linear Polarization (AoLP) modalities. The dataset consists of 302 training, 96 validation, and 102 testing samples, with each image having a resolution of 1224×1024. 

\textbf{MUSES}~\cite{muses} is a multimodal semantic segmentation dataset featuring synchronized RGB, event camera, and LiDAR data collected under diverse environmental conditions such as clear, foggy, rainy, and snowy weather, as well as different times of day (day and night). It provides high-resolution (1024×1024) images with pixel-level annotations for 11 semantic categories.

\textbf{DELIVER}~\cite{zhang2023delivering} is a large-scale multimodal semantic segmentation dataset constructed based on the CARLA simulator. It provides synchronized RGB images, depth maps, event data, LiDAR point clouds, and multi-view images organized as a panoramic cubemap. The dataset covers diverse environmental conditions, including clear, cloudy, foggy, rainy, and nighttime scenes, and introduces sensor failure cases such as motion blur, exposure issues, and LiDAR jitter. It contains 47,310 frames with semantic and instance annotations in 25 fine-grained categories. DELIVER is specifically designed to benchmark robust multimodal perception under challenging real-world conditions.

\textbf{Implementation details.} We train our model on eight NVIDIA GPUs with an initial learning rate (LR) of $6e^{-5}$, which is scheduled by the poly strategy with power 0.9 over 200 epochs. The first 10 epochs are to warm-up framework with 0.1× the original learning rate. We use AdamW as the optimizer, with epsilon of $1e^{-8}$, weight decay of $1e^{-2}$, and the batch size of 2 per GPU. The images are augmented by random resize with ratio 0.5–2.0, random horizontal flipping, random color jitter, random gaussian blur, and random cropping to 1024×1024 on DELIVER and MUSES, while to 512×512 on MCubeS. To ensure fair comparisons, all methods adopt the SegFormer-B0 backbone initialized with ImageNet-1K pre-trained weights.

\subsection{Experiment results}
\paragraph{Multimodal semantic segmentation.} 
Tab\ref{Efficiency} presents a comprehensive comparison between our model and various methods in terms of computational efficiency. 
Our method achieves remarkable efficiency, requiring only 1.90M parameters and 23.13 GFLOPs under the quad-modal setting (R-A-D-N), which corresponds to merely 18\% of the parameters and 50\% of the GFLOPs compared to CMNeXt. Despite this lightweight design, our method achieves comparable inference time and even higher FPS. While maintaining such a lightweight structure, our method still delivers excellent segmentation performance, as shown in Table~\ref{tab:deliver} and Table~\ref{tab:mcubes}. Specifically, on the DELIVER dataset, our model outperforms CMNeXt under most modality combinations and achieves the best mIoU of 60.55\% in the R-D-E setting. On the MCubeS dataset, our method consistently surpasses existing methods across various modality configurations, with particularly significant gains when all four modalities are fused, highlighting the strong effectiveness and generalization capability of our design.

\begin{wraptable}{r}{10cm}
\renewcommand{\tabcolsep}{1pt}
  \caption{Efficiency comparison on different methods.}
  \label{Efficiency}
  \centering
  \resizebox{\linewidth}{!}{
  \begin{tabular}{lcccccc}
   \toprule
   Method & Backbone & Modal     & \#Param (M) & GFLOPs & Inf. Time (ms) & FPS  \\
   \midrule
   CMX    & MiT-B0   & RGB       & 3.72        & 27.46 & \textbf{210} & \textbf{4.76}\\  
   CMNeXt & MiT-B0   & RGB       & 3.72        & 27.46 & \textbf{210} & \textbf{4.76}   \\
   StitchFusion & MiT-B0   & RGB & 4.26 & 29.37 & 244 & 4.10 \\
   Ours    & MiT-B0   & RGB       & \textbf{0.89}        & \textbf{20.17} & 229 & 4.37   \\
   \midrule
   CMX    & MiT-B0   & R-A     & 11.20       & 42.15 & 284 & 3.52   \\
   CMNeXt & MiT-B0   & R-A     & 10.29       & 43.44 & 380 & 2.63   \\
   StitchFusion & MiT-B0   & R-A & 4.40 & 39.12 & 267 & 3.74 \\
   Ours    & MiT-B0   & R-A     & \textbf{0.92}        & 2\textbf{0.72} & \textbf{247} & \textbf{4.05}   \\
   \midrule
   CMX    & MiT-B0   & R-A-D   & 11.25       & 43.72 & 371 & 2.69   \\
   CMNeXt & MiT-B0   & R-A-D   & 10.30       & 44.89 & 416 & 2.40   \\
   StitchFusion & MiT-B0   & R-A-D & 4.56 & 49.57 & 291 & 3.44 \\
   Ours    & MiT-B0   & R-A-D   & \textbf{1.06}        & \textbf{21.65} & \textbf{269} & \textbf{3.72}  \\
   \midrule
   CMX    & MiT-B0   & R-A-D-N & 11.28       & 45.20 & 378 & 2.65   \\
   CMNeXt & MiT-B0   & R-A-D-N & 10.30       & 46.29 & 424 & 2.36   \\
   StitchFusion & MiT-B0   & R-A-D-N & 4.77 & 60.71 & 335 & 2.99 \\
   Ours    & MiT-B0   & R-A-D-N & \textbf{1.90}        & \textbf{23.13} & \textbf{284} & \textbf{3.52} \\
    \bottomrule
  \end{tabular}}
\end{wraptable}

\begin{figure}[t!]
  \centering
  \includegraphics[width=0.85\linewidth]{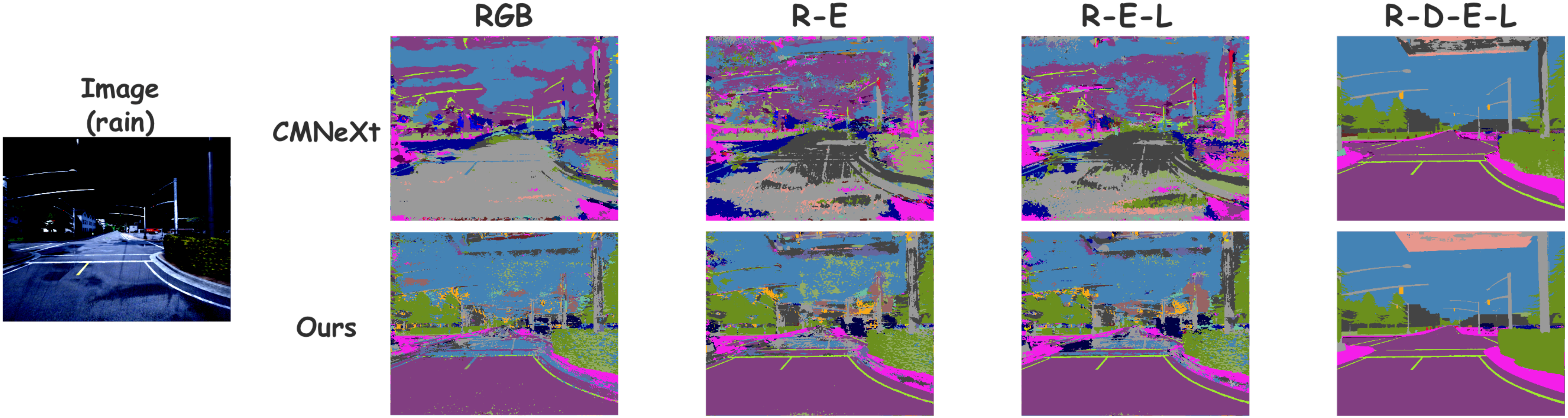}
  \caption{Visualization of networks trained with four modalities under missing modality scenarios and adverse weather conditions on the DELIVER dataset. More visualization refer to the appendix~\ref{val-full-deliver}.}
  \label{Visual}
\end{figure}

\paragraph{Robustness under arbitrary modality combinations.}
To evaluate the generalization capability of our framework, we adopt a train-once, test-everywhere setting. Specifically, our model is trained using all four modalities (RGB, Depth, Event, LiDAR), and evaluated on 15 different modality combinations, including all single-modal, dual-modal, tri-modal, and quad-modal configurations. This evaluation protocol reflects real-world deployment scenarios where sensor availability may vary dynamically. As shown in Table~\ref{tab:validation-b} and Table~\ref{tab:validation-a}, our model consistently outperforms the strong baseline CMNeXt under most combinations, despite being significantly lighter (1.9M vs. 34.8M parameters). Our model demonstrates strong performance in single-modal settings. For instance, it achieves substantial gains over CMNeXt on RGB-only and Depth-only inputs, highlighting its ability to extract meaningful representations from sparse or low-information inputs (e.g., 43.89 vs. 0.49 on Depth). Moreover, in dual-modal combinations such as RGB-Depth and Depth-LiDAR, our method continues to deliver robust performance. The significant margin on Depth-LiDAR (47.18 vs. 1.73) underscores the effectiveness of our dynamic fusion strategy, especially under challenging sparse-sparse fusion scenarios. Specifically, in higher-order fusion cases (tri-modal and quad-modal), our model matches or outperforms CMNeXt while using a fraction of the parameters. For example, we achieve 59.53 mIoU on the full-modality setting (RDEL) compared to CMNeXt’s 59.18, despite having 18\% parameters. Even under the competitive RDE configuration, our model maintains near-parity (58.80 vs. 59.03) with significantly lower computational cost. Figure~\ref{Visual} shows a visual comparison between our method and CMNeXt under rainy conditions using SegFormer-B0 as the backbone. All models are trained with four modalities (RGB, Depth, Event, and LiDAR), but tested under various missing modality scenarios. Our method demonstrates stronger robustness and semantic consistency, especially in adverse weather, preserving clearer object boundaries and road structure despite missing modalities.
\begin{table}[h!]
  \centering
  \begin{minipage}{0.48\textwidth}
    \centering
    \caption{Results on the DELIVER dataset}
    \label{tab:deliver}
    \renewcommand{\tabcolsep}{14pt}
  \resizebox{\linewidth}{!}{
    \begin{tabular}{llll}
      \toprule
      Method & Backbone & Modal & mIoU \\
      \midrule
      CMNeXt & MiT-B0 & RGB & \textbf{51.29} \\
      Ours   & MiT-B0 & RGB & 50.78 \\
      \midrule
      CMNeXt & MiT-B0 & R-D & 59.61 \\
      Ours   & MiT-B0 & R-D & \textbf{59.72} \\
      \midrule
      CMNeXt & MiT-B0 & R-D-E & 59.84 \\
      Ours   & MiT-B0 & R-D-E & \textbf{60.55} \\
      \midrule
      CMNeXt & MiT-B0 & R-D-E-L & 59.18 \\
      Ours   & MiT-B0 & R-D-E-L & \textbf{59.53} \\
      \bottomrule
    \end{tabular}}
  \end{minipage}%
  \hfill
  \begin{minipage}{0.48\textwidth}
    \centering
    \caption{Results on the MCubeS dataset}
    \label{tab:mcubes}
    \renewcommand{\tabcolsep}{14pt}
  \resizebox{\linewidth}{!}{
    \begin{tabular}{llll}
      \toprule
      Method & Backbone & Modal & mIoU \\
      \midrule
      CMNeXt & MiT-B0 & RGB & 24.96 \\
      Ours   & MiT-B0 & RGB & \textbf{40.55} \\
      \midrule
      CMNeXt & MiT-B0 & R-A & 37.21 \\
      Ours   & MiT-B0 & R-D & \textbf{43.46} \\
      \midrule
      CMNeXt & MiT-B0 & R-A-D & 38.72 \\
      Ours   & MiT-B0 & R-A-D & \textbf{41.34} \\
      \midrule
      CMNeXt & MiT-B0 & R-A-D-N & 36.16 \\
      Ours   & MiT-B0 & R-A-D-N & \textbf{43.40} \\
      \bottomrule
    \end{tabular}}
  \end{minipage}
  \vspace{-12pt}
\end{table}

\begin{table}[t!]
  \centering
  \caption{Validation with arbitrary modality combinations after four-modality training on MCubeS.}
    \renewcommand{\tabcolsep}{12pt}
  \label{tab:validation-b}
  \resizebox{\linewidth}{!}{
  \begin{tabular}{l|l|cccccccc}
    \toprule
    Method & Backbone & R & A & D & N & RA & RD & RN & AD \\
    \midrule
    CMNeXt & MiT-B0 & 2.53 & 1.71 & 1.28 & 0.23 & 6.47 & 7.73 & 7.56 & 4.13 \\
    \midrule
    Ours & MiT-B0 & \textbf{22.60} & \textbf{7.42} & \textbf{12.75} & \textbf{10.55} & \textbf{32.41} & \textbf{35.35} & \textbf{35.54} & \textbf{15.34} \\
    \midrule
    Method & Backbone & AN & DN & RAD & RAN & RDN & ADN & RADN & Mean \\
    \midrule
    CMNeXt & MiT-B0 & 3.76 & 0.36 & 5.37 & 5.58 & 9.14 & 4.56 & 36.16 & 6.44 \\
    \midrule
    Ours & MiT-B0 & \textbf{15.65} & \textbf{15.67} & \textbf{37.92} & \textbf{41.17} & \textbf{40.25} & \textbf{16.04} & \textbf{43.40} & \textbf{24.89} \\
    \bottomrule
  \end{tabular}}
  \vspace{-10pt}
\end{table}

\begin{table}[t!]
  \centering
  \caption{Validation with arbitrary modality combinations after four-modality training on DELIVER.}
  \label{tab:validation-a}
\renewcommand{\tabcolsep}{14pt}
\resizebox{\linewidth}{!}{
  \begin{tabular}{l|l|cccccccc}
    \toprule
    Method & Backbone & R & D & E & L & RD & RE & RL & DE \\
    \midrule
    CMNeXt & MiT-B0 & 0.86 & 0.49 & 0.66 & 0.37 & 47.06 & 9.97 & \textbf{13.75} & 2.63 \\
    \midrule
    Ours & MiT-B0 & \textbf{9.82} & \textbf{43.89} & \textbf{3.09} & \textbf{1.87} & \textbf{55.84} & \textbf{11.42} & 13.38 & \textbf{45.99} \\
    \midrule
    Method & Backbone & DL & EL & RDE & RDL & REL & DEL & RDEL & Mean \\
    \midrule
    CMNeXt & MiT-B0 & 1.73 & \textbf{2.85} & \textbf{59.03} & 59.18 & \textbf{14.73} & 39.07 & 59.18 & 20.77 \\
    \midrule
    Ours & MiT-B0 & \textbf{47.18} & 2.24 & 58.80 & \textbf{59.65} & 13.64 & \textbf{47.59} & \textbf{59.53} & \textbf{31.60} \\
    \bottomrule
  \end{tabular}}
  \vspace{-8pt}
\end{table}

\textbf{Synthetic-to-Real domain adaptation.}
To evaluate the generalization capability of our model in synthetic-to-real domain adaptation scenarios, we conduct experiments on the DELIVER and MUSES datasets. 
As shown in Table~\ref{Per-class mIoU results on DELIVER and MUSES dataset.}, we report the domain gap between source (DELIVER) and target (MUSES) domains for each semantic class. Our method consistently achieves significantly lower domain gaps compared to CMX and CMNeXt, with a mean gap of only 28.71\%, in contrast to 50.54\% for CMX and 58.39\% for CMNeXt. Notably, large reductions are observed in challenging categories such as Sky, Truck, and Vegetation. These results highlight the robustness and generalization capability of our modality-agnostic fusion framework under cross-domain settings, reinforcing its suitability for real-world deployment where target-domain supervision is scarce or unavailable.

\begin{table}[t!]
    \centering
    \caption{Per-class mIoU results on DELIVER and MUSES dataset, domain gaps means the performance gap between source and target domain. More results refer to the appendix~\ref{Supplementary per-class mIoU results on DELIVER and MUSES dataset.}}
    \label{Per-class mIoU results on DELIVER and MUSES dataset.}
    \scriptsize
    \renewcommand{\tabcolsep}{3pt}
    \resizebox{\textwidth}{!}{%
        \begin{tabular}{c|l|ccccccccccc|c}
            \toprule
                \textbf{Domain} & \textbf{Method} & \textbf{Build.} & \textbf{Bus} & \textbf{Fence} & \textbf{Pole} & \textbf{Road} & \textbf{Sidewalk} & \textbf{Sky} & \textbf{Terrain} & \textbf{Truck} & \textbf{Vege.} & \textbf{Wall} & \textbf{Mean} \\
                \midrule
\multirow{3}{*}{Domain Gaps$\downarrow$}
& CMX & 50.93 & 46.95 & 13.36 & 27.53 & 62.31 & 52.13 & 81.46 & 51.28 & 60.02 & 69.00 & 41.05 & 50.54 \\
& CMNeXt & 71.96 & 56.02 & 16.11 & 32.74 & 93.65 & 51.22 & 93.86 & 47.76 & 74.64 & 45.11 & 59.22 & 58.39 \\
 & Ours & 36.48 & 16.73 & 8.67 & 15.54 & 34.22 & 36.58 & 37.63 & 26.63 & 17.88 & 50.86 & 34.65 & 28.71 \\
                \bottomrule
            \end{tabular}
        }
        \vspace{-12pt}
\end{table}
\begin{figure}
  \centering
  \includegraphics[width=0.95\linewidth]{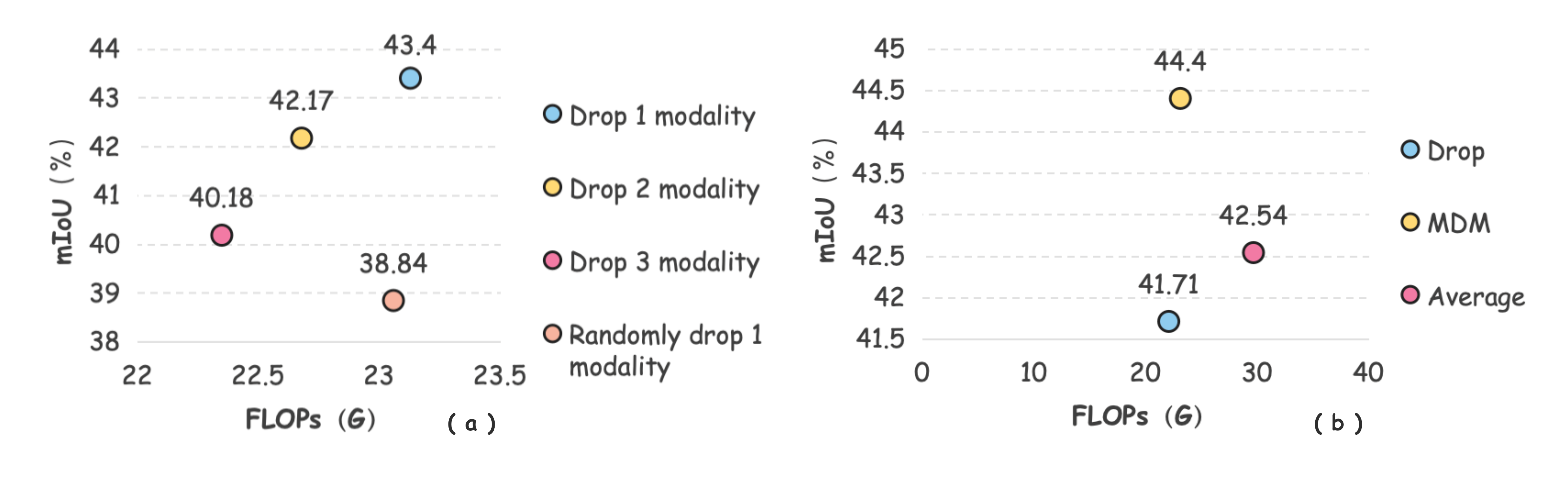}
  \caption{Trade-off Between mIoU and GFLOPs in MDM Ablation on the MCubeS Dataset.}
  \label{Trade-off Between mIoU and GFLOPs in MDM Ablation on the MCubeS Dataset.}
\end{figure}

\section{Ablation Study}
\label{sec:ablation study}
To further analyze the effectiveness of our modality-aware fusion strategy, we conduct extensive ablation studies on the MCubeS dataset using SegFormer-B0 as the backbone, aiming to explore the optimal Modality Dropping Module (MDM) strategies.  As shown in Figure~\ref{Trade-off Between mIoU and GFLOPs in MDM Ablation on the MCubeS Dataset.} (a), different modality dropping strategies reveal a clear trade-off between segmentation performance and computational efficiency. Specifically, dropping a single modality per stage achieves the best result with 43.4\% mIoU at 23.13 GFLOPs. This suggests that while partial modality reduction reduces redundancy, it still retains sufficient multimodal information for robust learning. In contrast, dropping two or three modalities leads to a significant performance drop (42.17\% and 40.18\% mIoU respectively), indicating that excessive pruning harms semantic completeness. Notably, randomly dropping one modality yields the lowest performance (38.84\% mIoU), likely due to the unpredictability disrupting feature consistency, which highlights the importance of stable fusion pathways. These results demonstrate a clear trade-off: reducing modality redundancy improves efficiency but may hurt segmentation performance. Our design allows for flexible control based on deployment constraints.

Complementing this analysis, we further compare different modality selection mechanisms in Figure~\ref{Trade-off Between mIoU and GFLOPs in MDM Ablation on the MCubeS Dataset.} (b). Naïvely dropping the weakest modality reduces computation to 21.2 GFLOPs but sacrifices accuracy (41.71\% mIoU), implying that even weaker modalities may carry task-relevant cues. Average fusion marginally improves performance (42.54\% mIoU), but its high computational cost (29.5 GFLOPs) reflects inefficient use of redundant modalities. In contrast, our proposed Modality Dropping Module (MDM) dynamically selects and compresses features from weaker modalities before discarding noisy signals, achieving the best balance of 44.4\% mIoU with only 23.1 GFLOPs. This demonstrates that intelligent modality selection not only preserves informative features but also avoids unnecessary overhead.

\section{Conclusion and Limitations}
\label{sec:conclusion and limitations}
In this work, we presented EGFormer, a lightweight and generalize multimodal semantic segmentation framework that addresses both accuracy and efficiency. By introducing the Any-modal Scoring Module (ASM) and the Modal Dropping Module (MDM), EGFormer dynamically evaluates and selects informative modalities during inference, leading to significant reductions in model size and computational cost. Extensive experiments on various datasets confirmed that EGFormer maintains competitive segmentation performance while reducing parameters by up to 88\% and GFLOPs by 50\%. Furthermore, EGFormer exhibits strong generalization ability in challenging synthetic-to-real scenarios and achieves state-of-the-art results under unsupervised domain adaptation settings. These results validate the practical value of EGFormer in real-world multimodal applications, especially where computational resources or input modalities are constrained.

\textbf{Limitations.} While our model demonstrates a reduced domain gap in the Synthetic-to-Real domain adaptation task compared to several state-of-the-art methods, there remains room for further optimization. In future work, we plan to focus on enhancing the model’s transferability and domain adaptation performance through more advanced adaptation techniques.

\section*{Broader Impacts}
\label{sec:broaderimpacts}
This work presents a lightweight, modality-aware segmentation framework tailored for resource-constrained environments. By selectively leveraging relevant sensor modalities and discarding redundant or noisy ones, the model delivers competitive performance with reduced computational overhead. This makes it well-suited for applications such as mobile robotics, drones, wearable devices, and edge-based infrastructure where energy, memory, or processing power is limited. 
Ethically, robust performance in degraded environments promotes accessibility and equity, enabling AI systems to operate beyond well-resourced settings. However, care must be taken to avoid deprioritizing sensors critical to minority use cases. We encourage fairness evaluations across environmental contexts and the use of interpretability tools to support transparent, responsible deployment.

\clearpage
\bibliographystyle{ieeetr}  
\bibliography{reference}

\begin{thebibliography}{10}

\bibitem{alonso2019evsegnet}
I.~Alonso and A.~C. Murillo, ``Ev-segnet: Semantic segmentation for event-based cameras,'' in {\em Proceedings of the IEEE/CVF Conference on Computer Vision and Pattern Recognition Workshops (CVPRW)}, pp.~0--0, IEEE, 2019.

\bibitem{cao2021shapeconv}
J.~Cao, H.~Leng, D.~Lischinski, {\em et~al.}, ``Shapeconv: Shape-aware convolutional layer for indoor rgb-d semantic segmentation,'' in {\em Proceedings of the IEEE/CVF International Conference on Computer Vision (ICCV)}, (Montreal, Canada), pp.~7088--7097, IEEE, 2021.

\bibitem{chen2020bidirectional}
X.~Chen, K.~Lin, J.~Wang, {\em et~al.}, ``Bi-directional cross-modality feature propagation with separation-and-aggregation gate for rgb-d semantic segmentation,'' in {\em Proceedings of the European Conference on Computer Vision (ECCV)}, (Cham), pp.~561--577, Springer, 2020.

\bibitem{jia2023event}
Z.~Jia, K.~You, W.~He, {\em et~al.}, ``Event-based semantic segmentation with posterior attention,'' {\em IEEE Transactions on Image Processing}, vol.~32, pp.~1829--1842, 2023.

\bibitem{sun2019rtfnet}
Y.~Sun, W.~Zuo, and M.~Liu, ``Rtfnet: Rgb-thermal fusion network for semantic segmentation of urban scenes,'' {\em IEEE Robotics and Automation Letters}, vol.~4, no.~3, pp.~2576--2583, 2019.

\bibitem{zhang2023delivering}
J.~Zhang, R.~Liu, H.~Shi, {\em et~al.}, ``Delivering arbitrary-modal semantic segmentation,'' in {\em Proceedings of the IEEE/CVF Conference on Computer Vision and Pattern Recognition (CVPR)}, (New Orleans, LA), pp.~1136--1147, IEEE, 2023.

\bibitem{lyu2024omnibind}
Y.~Lyu, X.~Zheng, D.~Kim, and L.~Wang, ``Omnibind: Teach to build unequal-scale modality interaction for omni-bind of all,'' {\em arXiv preprint arXiv:2405.16108}, 2024.

\bibitem{hazirbas2016fusenet}
C.~Hazirbas, L.~Ma, C.~Domokos, and D.~Cremers, ``Fusenet: Incorporating depth into semantic segmentation via fusion-based cnn architecture,'' in {\em Proceedings of the Asian Conference on Computer Vision (ACCV)}, (Cham), pp.~213--228, Springer, 2016.

\bibitem{elmadawi2019rgb}
K.~El~Madawi, H.~Rashed, A.~El~Sallab, {\em et~al.}, ``Rgb and lidar fusion based 3d semantic segmentation for autonomous driving,'' in {\em Proceedings of the 2019 IEEE Intelligent Transportation Systems Conference (ITSC)}, pp.~7--12, IEEE, 2019.

\bibitem{li2023mseg3d}
J.~Li, H.~Dai, H.~Han, {\em et~al.}, ``Mseg3d: Multi-modal 3d semantic segmentation for autonomous driving,'' in {\em Proceedings of the IEEE/CVF Conference on Computer Vision and Pattern Recognition (CVPR)}, pp.~21694--21704, IEEE, 2023.

\bibitem{lyu2024unibind}
Y.~Lyu, X.~Zheng, J.~Zhou, and L.~Wang, ``Unibind: Llm-augmented unified and balanced representation space to bind them all,'' in {\em Proceedings of the IEEE/CVF Conference on Computer Vision and Pattern Recognition}, pp.~26752--26762, 2024.

\bibitem{yao2024sam}
B.~Yao, Y.~Deng, Y.~Liu, {\em et~al.}, ``Sam-event-adapter: Adapting segment anything model for event-rgb semantic segmentation,'' in {\em Proceedings of the 2024 IEEE International Conference on Robotics and Automation (ICRA)}, pp.~9093--9100, IEEE, 2024.

\bibitem{zhao2025eseg}
Y.~Zhao, G.~Lyu, K.~Li, {\em et~al.}, ``Eseg: Event-based segmentation boosted by explicit edge-semantic guidance,'' in {\em Proceedings of the AAAI Conference on Artificial Intelligence}, vol.~39, pp.~10510--10518, 2025.

\bibitem{zheng2024eventdance}
X.~Zheng and L.~Wang, ``Eventdance: Unsupervised source-free cross-modal adaptation for event-based object recognition,'' in {\em Proceedings of the IEEE/CVF Conference on Computer Vision and Pattern Recognition}, pp.~17448--17458, 2024.

\bibitem{zhou2024exact}
J.~Zhou, X.~Zheng, Y.~Lyu, and L.~Wang, ``Exact: Language-guided conceptual reasoning and uncertainty estimation for event-based action recognition and more,'' in {\em Proceedings of the IEEE/CVF Conference on Computer Vision and Pattern Recognition}, pp.~18633--18643, 2024.

\bibitem{brodermann2025cafuser}
T.~Brödermann, C.~Sakaridis, Y.~Fu, {\em et~al.}, ``Cafuser: Condition-aware multimodal fusion for robust semantic perception of driving scenes,'' {\em IEEE Robotics and Automation Letters}, 2025.

\bibitem{li2024stitchfusion}
B.~Li, D.~Zhang, Z.~Zhao, {\em et~al.}, ``Stitchfusion: Weaving any visual modalities to enhance multimodal semantic segmentation,'' {\em arXiv preprint arXiv:2408.01343}, 2024.

\bibitem{zheng2024centering}
X.~Zheng, Y.~Lyu, J.~Zhou, {\em et~al.}, ``Centering the value of every modality: Towards efficient and resilient modality-agnostic semantic segmentation,'' in {\em Proceedings of the European Conference on Computer Vision (ECCV)}, (Cham), pp.~192--212, Springer Nature Switzerland, 2024.

\bibitem{wang2022multimodal}
Y.~Wang, X.~Chen, L.~Cao, {\em et~al.}, ``Multimodal token fusion for vision transformers,'' in {\em Proceedings of the IEEE/CVF Conference on Computer Vision and Pattern Recognition (CVPR)}, pp.~12186--12195, IEEE, 2022.

\bibitem{zheng2024360sfuda++}
X.~Zheng, P.~Y. Zhou, A.~V. Vasilakos, and L.~Wang, ``360sfuda++: Towards source-free uda for panoramic segmentation by learning reliable category prototypes,'' {\em IEEE Transactions on Pattern Analysis and Machine Intelligence}, 2024.

\bibitem{zheng2024semantics}
X.~Zheng, P.~Zhou, A.~V. Vasilakos, and L.~Wang, ``Semantics distortion and style matter: Towards source-free uda for panoramic segmentation,'' in {\em Proceedings of the IEEE/CVF Conference on Computer Vision and Pattern Recognition}, pp.~27885--27895, 2024.

\bibitem{zheng2023both}
X.~Zheng, J.~Zhu, Y.~Liu, Z.~Cao, C.~Fu, and L.~Wang, ``Both style and distortion matter: Dual-path unsupervised domain adaptation for panoramic semantic segmentation,'' in {\em Proceedings of the IEEE/CVF Conference on Computer Vision and Pattern Recognition}, pp.~1285--1295, 2023.

\bibitem{zheng2023look}
X.~Zheng, T.~Pan, Y.~Luo, and L.~Wang, ``Look at the neighbor: Distortion-aware unsupervised domain adaptation for panoramic semantic segmentation,'' in {\em Proceedings of the IEEE/CVF ICCV}, pp.~18687--18698, 2023.

\bibitem{liang2022multimodal}
Y.~Liang, R.~Wakaki, S.~Nobuhara, {\em et~al.}, ``Multimodal material segmentation,'' in {\em Proceedings of the IEEE/CVF Conference on Computer Vision and Pattern Recognition (CVPR)}, pp.~19800--19808, IEEE, 2022.

\bibitem{muses}
X.~Liu, Y.~Hu, S.~Bai, {\em et~al.}, ``Multi-shot temporal event localization: A benchmark,'' in {\em Proceedings of the IEEE/CVF Conference on Computer Vision and Pattern Recognition (CVPR)}, pp.~12596--12606, IEEE, 2021.

\bibitem{audebert2018beyond}
N.~Audebert, B.~Le~Saux, and S.~Lefèvre, ``Beyond rgb: Very high resolution urban remote sensing with multimodal deep networks,'' {\em ISPRS Journal of Photogrammetry and Remote Sensing}, vol.~140, pp.~20--32, 2018.

\bibitem{dimitrovski2024unet}
I.~Dimitrovski, V.~Spasev, S.~Loshkovska, {\em et~al.}, ``U-net ensemble for enhanced semantic segmentation in remote sensing imagery,'' {\em Remote Sensing}, vol.~16, no.~12, p.~2077, 2024.

\bibitem{wang2024metasegnet}
L.~Wang, S.~Dong, Y.~Chen, {\em et~al.}, ``Metasegnet: Metadata-collaborative vision-language representation learning for semantic segmentation of remote sensing images,'' {\em IEEE Transactions on Geoscience and Remote Sensing}, 2024.

\bibitem{zheng2024transformer}
X.~Zheng, Y.~Luo, C.~Fu, K.~Liu, and L.~Wang, ``Transformer-cnn cohort: Semi-supervised semantic segmentation by the best of both students,'' in {\em 2024 IEEE International Conference on Robotics and Automation (ICRA)}, pp.~11147--11154, IEEE, 2024.

\bibitem{DBLP:journals/pr/ZhengLZW25}
X.~Zheng, Y.~Luo, P.~Zhou, and L.~Wang, ``Distilling efficient vision transformers from cnns for semantic segmentation,'' {\em Pattern Recognit.}, vol.~158, p.~111029, 2025.

\bibitem{DBLP:journals/pr/ChenDZZM24}
J.~Chen, D.~Deguchi, C.~Zhang, X.~Zheng, and H.~Murase, ``Frozen is better than learning: {A} new design of prototype-based classifier for semantic segmentation,'' {\em Pattern Recognit.}, vol.~152, p.~110431, 2024.

\bibitem{munasinghe2022covered}
C.~Munasinghe, F.~M. Amin, D.~Scaramuzza, {\em et~al.}, ``Covered: Collaborative robot environment dataset for 3d semantic segmentation,'' in {\em Proceedings of the 2022 IEEE 27th International Conference on Emerging Technologies and Factory Automation (ETFA)}, pp.~1--4, IEEE, 2022.

\bibitem{long2015fully}
J.~Long, E.~Shelhamer, and T.~Darrell, ``Fully convolutional networks for semantic segmentation,'' in {\em Proceedings of the IEEE Conference on Computer Vision and Pattern Recognition (CVPR)}, pp.~3431--3440, IEEE, 2015.

\bibitem{chen2017deeplab}
L.-C. Chen, G.~Papandreou, I.~Kokkinos, K.~Murphy, and A.~L. Yuille, ``Deeplab: Semantic image segmentation with deep convolutional nets, atrous convolution, and fully connected crfs,'' {\em IEEE Transactions on Pattern Analysis and Machine Intelligence}, vol.~40, no.~4, pp.~834--848, 2018.

\bibitem{chen2018encoder}
L.-C. Chen, Y.~Zhu, G.~Papandreou, F.~Schroff, and H.~Adam, ``Encoder-decoder with atrous separable convolution for semantic image segmentation,'' in {\em Proceedings of the European Conference on Computer Vision (ECCV)}, pp.~801--818, Springer, 2018.

\bibitem{hou2020strip}
Q.~Hou, L.~Zhang, M.-M. Cheng, {\em et~al.}, ``Strip pooling: Rethinking spatial pooling for scene parsing,'' in {\em Proceedings of the IEEE/CVF Conference on Computer Vision and Pattern Recognition (CVPR)}, pp.~4003--4012, IEEE, 2020.

\bibitem{zhao2017pyramid}
H.~Zhao, J.~Shi, X.~Qi, X.~Wang, and J.~Jia, ``Pyramid scene parsing network,'' in {\em Proceedings of the IEEE Conference on Computer Vision and Pattern Recognition (CVPR)}, pp.~2881--2890, IEEE, 2017.

\bibitem{choi2020cars}
S.~Choi, J.~T. Kim, and J.~Choo, ``Cars can't fly up in the sky: Improving urban-scene segmentation via height-driven attention networks,'' in {\em Proceedings of the IEEE/CVF Conference on Computer Vision and Pattern Recognition (CVPR)}, pp.~9373--9382, IEEE, 2020.

\bibitem{fu2019dual}
J.~Fu, J.~Liu, H.~Tian, Y.~Li, Y.~Bao, Z.~Fang, and H.~Lu, ``Dual attention network for scene segmentation,'' in {\em Proceedings of the IEEE/CVF Conference on Computer Vision and Pattern Recognition (CVPR)}, pp.~3146--3154, IEEE, 2019.

\bibitem{huang2019ccnet}
Z.~Huang, X.~Wang, L.~Huang, C.~Huang, Y.~Wei, and W.~Liu, ``Ccnet: Criss-cross attention for semantic segmentation,'' in {\em Proceedings of the IEEE/CVF International Conference on Computer Vision (ICCV)}, pp.~603--612, IEEE, 2019.

\bibitem{yuan2021ocnet}
Y.~Yuan, L.~Huang, J.~Guo, C.~Zhang, X.~Chen, and J.~Wang, ``Ocnet: Object context for semantic segmentation,'' {\em International Journal of Computer Vision (IJCV)}, vol.~129, no.~5, pp.~1106--1121, 2021.

\bibitem{borse2021inverseform}
S.~Borse, Y.~Wang, Y.~Zhang, and F.~Porikli, ``Inverseform: A loss function for structured boundary-aware segmentation,'' in {\em Proceedings of the IEEE/CVF Conference on Computer Vision and Pattern Recognition (CVPR)}, pp.~5901--5911, IEEE, 2021.

\bibitem{ding2019boundary}
H.~Ding, X.~Jiang, A.~Q. Liu, N.~M. Thalmann, and G.~Wang, ``Boundary-aware feature propagation for scene segmentation,'' in {\em Proceedings of the IEEE/CVF International Conference on Computer Vision (ICCV)}, pp.~6819--6829, IEEE, 2019.

\bibitem{li2020improving}
X.~Li, X.~Li, L.~Zhang, G.~Cheng, J.~Shi, Z.~Lin, S.~Tan, and Y.~Tong, ``Improving semantic segmentation via decoupled body and edge supervision,'' in {\em Computer Vision -- ECCV 2020: 16th European Conference, Proceedings, Part XVII}, (Glasgow, UK), pp.~435--452, Springer, 2020.

\bibitem{ranftl2021vision}
R.~Ranftl, A.~Bochkovskiy, and V.~Koltun, ``Vision transformers for dense prediction without convolutions,'' {\em arXiv preprint arXiv:2106.06195}, 2021.

\bibitem{gu2022multi}
J.~Gu, H.~Kwon, D.~Wang, W.~Ye, M.~Li, Y.-H. Chen, L.~Lai, V.~Chandra, and D.~Z. Pan, ``Multi-scale high-resolution vision transformer for semantic segmentation,'' in {\em Proceedings of the IEEE/CVF Conference on Computer Vision and Pattern Recognition (CVPR)}, pp.~16256--16266, IEEE, 2022.

\bibitem{guo2022segnext}
M.-H. Guo, C.-Z. Lu, Q.~Hou, Z.~Liu, M.-M. Cheng, and S.-M. Hu, ``Segnext: Rethinking convolutional attention design for semantic segmentation,'' in {\em Advances in Neural Information Processing Systems (NeurIPS)}, 2022.

\bibitem{liu2022swin}
Z.~Liu, H.~Hu, Y.~Lin, Z.~Yao, Z.~Xie, Y.~Wei, J.~Ning, Y.~Cao, Z.~Zhang, L.~Dong, {\em et~al.}, ``Swin transformer v2: Scaling up capacity and resolution,'' in {\em Proceedings of the IEEE/CVF Conference on Computer Vision and Pattern Recognition (CVPR)}, pp.~12009--12019, IEEE, 2022.

\bibitem{liu2021swin}
Z.~Liu, Y.~Lin, Y.~Cao, H.~Hu, Y.~Wei, Z.~Zhang, S.~Lin, and B.~Guo, ``Swin transformer: Hierarchical vision transformer using shifted windows,'' in {\em Proceedings of the IEEE/CVF International Conference on Computer Vision (ICCV)}, pp.~10012--10022, IEEE, 2021.

\bibitem{strudel2021segmenter}
R.~Strudel, R.~Garcia, I.~Laptev, and C.~Schmid, ``Segmenter: Transformer for semantic segmentation,'' in {\em Proceedings of the IEEE/CVF International Conference on Computer Vision (ICCV)}, pp.~7262--7272, IEEE, 2021.

\bibitem{xie2021segformer}
E.~Xie, W.~Wang, Z.~Yu, A.~Anandkumar, J.~M. Alvarez, and P.~Luo, ``Segformer: Simple and efficient design for semantic segmentation with transformers,'' {\em Advances in Neural Information Processing Systems (NeurIPS)}, vol.~34, pp.~12077--12090, 2021.

\bibitem{zhang2022segvit}
B.~Zhang, Z.~Tian, Q.~Tang, X.~Chu, X.~Wei, C.~Shen, and Y.~Liu, ``Segvit: Semantic segmentation with plain vision transformers,'' in {\em Advances in Neural Information Processing Systems (NeurIPS)}, 2022.

\bibitem{zheng2021rethinking}
S.~Zheng, J.~Lu, H.~Zhao, X.~Zhu, Z.~Luo, Y.~Wang, Y.~Fu, J.~Feng, T.~Xiang, P.~H.~S. Torr, and L.~Zhang, ``Rethinking semantic segmentation from a sequence-to-sequence perspective with transformers,'' in {\em Proceedings of the IEEE/CVF Conference on Computer Vision and Pattern Recognition (CVPR)}, pp.~6881--6890, IEEE, 2021.

\bibitem{borse2023xalign}
S.~Borse, M.~Klingner, V.~R. Kumar, H.~Cai, A.~Almuzairee, S.~Yogamani, and F.~Porikli, ``X-align: Cross-modal cross-view alignment for bird’s-eye-view segmentation,'' in {\em Proceedings of the IEEE/CVF Winter Conference on Applications of Computer Vision (WACV)}, pp.~3287--3297, IEEE, 2023.

\bibitem{chen2022modality}
G.~Chen, F.~Shao, X.~Chai, H.~Chen, Q.~Jiang, X.~Meng, and Y.-S. Ho, ``Modality-induced transfer-fusion network for rgb-d and rgb-t salient object detection,'' {\em IEEE Transactions on Circuits and Systems for Video Technology}, vol.~33, no.~4, pp.~1787--1801, 2022.

\bibitem{hui2023bridging}
T.~Hui, Z.~Xun, F.~Peng, J.~Huang, X.~Wei, X.~Wei, J.~Dai, J.~Han, and S.~Liu, ``Bridging search region interaction with template for rgb-t tracking,'' in {\em Proceedings of the IEEE/CVF Conference on Computer Vision and Pattern Recognition (CVPR)}, pp.~13630--13639, IEEE, 2023.

\bibitem{liao2022cross}
G.~Liao, W.~Gao, G.~Li, J.~Wang, and S.~Kwong, ``Cross-collaborative fusion-encoder network for robust rgb-thermal salient object detection,'' {\em IEEE Transactions on Circuits and Systems for Video Technology}, vol.~32, no.~11, pp.~7646--7661, 2022.

\bibitem{mei2022glass}
H.~Mei, B.~Dong, W.~Dong, J.~Yang, S.-H. Baek, F.~Heide, P.~Peers, X.~Wei, and X.~Yang, ``Glass segmentation using intensity and spectral polarization cues,'' in {\em Proceedings of the IEEE/CVF Conference on Computer Vision and Pattern Recognition (CVPR)}, pp.~12622--12631, IEEE, 2022.

\bibitem{pang2023caver}
Y.~Pang, X.~Zhao, L.~Zhang, and H.~Lu, ``Caver: Cross-modal view-mixed transformer for bi-modal salient object detection,'' {\em IEEE Transactions on Image Processing}, vol.~32, pp.~892--904, 2023.

\bibitem{liao2025benchmarking}
C.~Liao, K.~Lei, X.~Zheng, J.~Moon, Z.~Wang, Y.~Wang, D.~P. Paudel, L.~Van~Gool, and X.~Hu, ``Benchmarking multi-modal semantic segmentation under sensor failures: Missing and noisy modality robustness,'' {\em arXiv preprint arXiv:2503.18445}, 2025.

\bibitem{zheng2025reducing}
X.~Zheng, Y.~Lyu, L.~Jiang, D.~P. Paudel, L.~Van~Gool, and X.~Hu, ``Reducing unimodal bias in multi-modal semantic segmentation with multi-scale functional entropy regularization,'' {\em arXiv preprint arXiv:2505.06635}, 2025.

\bibitem{liao2025memorysam}
C.~Liao, X.~Zheng, Y.~Lyu, H.~Xue, Y.~Cao, J.~Wang, K.~Yang, and X.~Hu, ``Memorysam: Memorize modalities and semantics with segment anything model 2 for multi-modal semantic segmentation,'' {\em arXiv preprint arXiv:2503.06700}, 2025.

\bibitem{zhao2025unveiling}
J.~Zhao, F.~Teng, K.~Luo, G.~Zhao, Z.~Li, X.~Zheng, and K.~Yang, ``Unveiling the potential of segment anything model 2 for rgb-thermal semantic segmentation with language guidance,'' {\em arXiv preprint arXiv:2503.02581}, 2025.

\bibitem{zheng2024learning}
X.~Zheng, Y.~Lyu, and L.~Wang, ``Learning modality-agnostic representation for semantic segmentation from any modalities,'' in {\em European Conference on Computer Vision}, pp.~146--165, Springer, 2024.

\bibitem{zheng2024magic++}
X.~Zheng, Y.~Lyu, L.~Jiang, J.~Zhou, L.~Wang, and X.~Hu, ``Magic++: Efficient and resilient modality-agnostic semantic segmentation via hierarchical modality selection,'' {\em arXiv preprint arXiv:2412.16876}, 2024.

\bibitem{zhu2024customize}
C.~Zhu, B.~Xiao, L.~Shi, S.~Xu, and X.~Zheng, ``Customize segment anything model for multi-modal semantic segmentation with mixture of lora experts,'' {\em arXiv preprint arXiv:2412.04220}, 2024.

\bibitem{zheng2023deep}
X.~Zheng, Y.~Liu, Y.~Lu, T.~Hua, T.~Pan, W.~Zhang, D.~Tao, and L.~Wang, ``Deep learning for event-based vision: A comprehensive survey and benchmarks,'' {\em arXiv preprint arXiv:2302.08890}, 2023.

\end{thebibliography}

\clearpage
\appendix

\section{Technical Appendices and Supplementary Material}
In this Appendices, we provide a comprehensive set of visualizations and analyses to elucidate the performance and capabilities of our EGFormer method. Table~\ref{Full results on DELIVER} provides a comprehensive evaluation of our proposed method using Segformer-B0 as the pretrained backbone on the DELIVER dataset. The table reports the mIoU results across all individual modalities (R: RGB, D: Depth, E: Event, L: LiDAR) and their combinations. We assess performance on both single-modal and multi-modal configurations, including pairs (e.g., RD, RE), triplets (e.g., RDE, RDL), and the full combination RDEL. Among single-modal inputs, depth (D: 54.46) and RGB (R: 50.78) yield the best results individually. In dual-modal combinations, RD (59.72) and DE (54.07) provide strong performance gains over single modalities. For triple-modal settings, RDL (59.92) and RDEL (59.53) deliver high mIoU scores, demonstrating the benefits of rich multimodal fusion. The best mean performance (49.47) is observed with the full combination of RDEL, indicating the advantage of using all available modalities. These results highlight the effectiveness of the EGFormer in learning robust representations from heterogeneous sensor inputs.
\begin{table}[H]
  \centering
  \caption{Full results on DELIVER}
  \label{Full results on DELIVER}
  \begin{tabular}{l|l|cccccccc}
    \toprule
    Method & Backbone & R & D & E & L & RD & RE & RL & DE \\
    \midrule
    Ours & MiT-B0 & 50.78 & 54.46 & 26.74 & 29.20 & 59.72 & 50.03 & 50.18 & 54.07 \\
    \midrule
    Method & Backbone & DL & EL & RDE & RDL & REL & DEL & RDEL & Mean \\
    \midrule
    Ours & MiT-B0 & 53.14 & 32.12 & 60.55 & 59.92 & 48.73 & 52.89 & 59.53 & 49.47 \\
    \bottomrule
  \end{tabular}
\end{table}
Table~\ref{Full results on MCubeS} presents the detailed performance of our EGFormer method on the MCubeS dataset, again using Segformer-B0 as the pretrained backbone. In line with the DELIVER evaluation, we assess all individual modalities (R: RGB, A: AoLP, D: DoLP, N: NIR) and their combinations. Among the single-modal inputs, RGB (R: 40.55) and NIR (N: 37.20) yield the highest mIoU scores. For dual-modal configurations, RA (43.46), RD (42.22), and RN (41.01) all demonstrate significant improvements over single-modality baselines, confirming the benefits of multimodal fusion. In higher-order combinations, such as RAD (41.34), RAN (41.18), and RADN (43.40), the model further capitalizes on the complementary nature of the inputs. Notably, RADN achieves the highest mIoU score (43.40) among all configurations, highlighting the effectiveness of full-modal fusion in the MCubeS setting. These consistent gains across modality combinations validate the robustness and adaptability of our EGFormer in addressing challenging multimodal segmentation tasks.

\begin{table}[H]
  \centering
  \caption{Full results on MCubeS}
  \label{Full results on MCubeS}
  \begin{tabular}{l|l|cccccccc}
    \toprule
    Method & Backbone & R & A & D & N & RA & RD & RN & AD \\
    \midrule
    Ours & MiT-B0 & 40.55 & 29.36 & 29.83 & 37.20 & 43.46 & 42.22 & 41.01 & 33.37 \\
    \midrule
    Method & Backbone & AN & DN & RAD & RAN & RDN & ADN & RADN & Mean \\
    \midrule
    Ours & MiT-B0 & 34.63 & 35.88 & 41.34 & 41.18 & 40.06 & 37.44 & 43.40 & 30.06 \\
    \bottomrule
  \end{tabular}
\end{table}

Table~\ref{Supplementary per-class mIoU results on DELIVER and MUSES dataset.} presents the per-class mIoU performance on the DELIVER (source) and MUSES (target) datasets, as well as the corresponding domain gaps—defined as the difference in performance between the source and target domains. We compare three methods: CMX, CMNeXt, and our proposed model (Ours), with Ours 1 representing the model pretrained on DELIVER and Ours 2 being further fine-tuned on MUSES. Compared to CMX and CMNeXt, our model achieves consistently lower domain gaps across most categories. For instance, Ours reduces the domain gap in challenging classes such as Fence (from 13.36/16.11 to 8.67), Sky (from 81.46/93.86 to 37.63), and Sidewalk (from 52.13/51.22 to 36.58), indicating improved cross-domain generalization. The mean domain gap of our method (28.71) is significantly lower than that of CMX (50.54) and CMNeXt (55.32), demonstrating the effectiveness of our approach in reducing performance degradation under domain shift. Moreover, the use of MUSES fine-tuning (Ours 2) further improves source-side accuracy in certain categories (e.g., Pole, Road, Wall), while still maintaining reasonable generalization performance on the target domain. This highlights the benefit of adapting to the target distribution while preserving cross-modal robustness.
\begin{table}[t!]
    \centering
    \caption{Supplementary per-class mIoU results on DELIVER and MUSES dataset, domain gaps refers to the performance gap between source and target domain. Ours 1 is pretrained on DELIVER, and Our 2 is fine-tuned on MUSES.}
    \label{Supplementary per-class mIoU results on DELIVER and MUSES dataset.}
    \scriptsize
    \renewcommand{\tabcolsep}{3pt}
    \resizebox{\textwidth}{!}{%
        \begin{tabular}{c|l|ccccccccccc|c}
            \toprule
                \textbf{Domain} & \textbf{Method} & \textbf{Build.} & \textbf{Bus} & \textbf{Fence} & \textbf{Pole} & \textbf{Road} & \textbf{Sidewalk} & \textbf{Sky} & \textbf{Terrain} & \textbf{Truck} & \textbf{Vege.} & \textbf{Wall} & \textbf{Mean} \\
                \midrule
 \multirow{3}{*}{Source} 
 & CMX & 71.53 & 46.95 & 13.36 & 30.71 & 96.53 & 52.73 & 96.46 & 51.28 & 60.02 & 69.00 & 43.30 &57.44\\
 & CMNeXt & 71.96 & 56.02 & 17.45 & 32.74 & 96.96 & 55.28 & 96.86 & 53.33 & 74.68 & 71.83 & 59.22 & 62.39\\
 & Ours 1 & 65.88 &46.62 & 23.67 & 31.15 & 97.62 & 46.86 & 59.21 & 45.70 & 53.63 & 73.70 & 44.74 & 53.53 \\
 & Ours 2 & 55.99 & 16.73 & 13.78 & 21.26 & 94.92 & 36.97 & 94.09 & 35.81 & 17.88 & 63.81 & 34.85 & 44.19 \\
 \midrule
 \multirow{3}{*}{Target}
 & CMX & 20.60 & 0.00 & 0.00 & 3.18 & 34.22 & 0.60 & 15.00 & 0.00 & 0.00 & 0.00 & 2.25 & 6.90\\
  & CMNeXt & 0.00 & 0.00 & 1.34 & 0.00 & 3.31 & 4.06 & 3.00 & 5.57 & 0.04 & 26.72 & 0.00 & 4.00 \\
 & Ours 
 & 19.51 & 0.00 & 5.11 & 5.72 & 60.70 & 0.39 & 56.46 & 9.18 & 0.00 & 12.95 & 0.20 & 15.48 \\
 \midrule
\multirow{3}{*}{Domain Gaps$\downarrow$}
& CMX & 50.93 & 46.95 & 13.36 & 27.53 & 62.31 & 52.13 & 81.46 & 51.28 & 60.02 & 69.00 & 41.05 & 50.54 \\
& CMNeXt & 71.96 & 56.02 & 16.11 & 32.74 & 93.65 & 51.22 & 93.86 & 47.76 & 74.64 & 45.11 & 59.22 & 58.39 \\
 & Ours & 36.48 & 16.73 & 8.67 & 15.54 & 34.22 & 36.58 & 37.63 & 26.63 & 17.88 & 50.86 & 34.65 & 28.71 \\
                \bottomrule
            \end{tabular}
        }
        \vspace{-12pt}
\end{table}

\begin{figure}[t!]
  \centering
  \includegraphics[width=\linewidth]{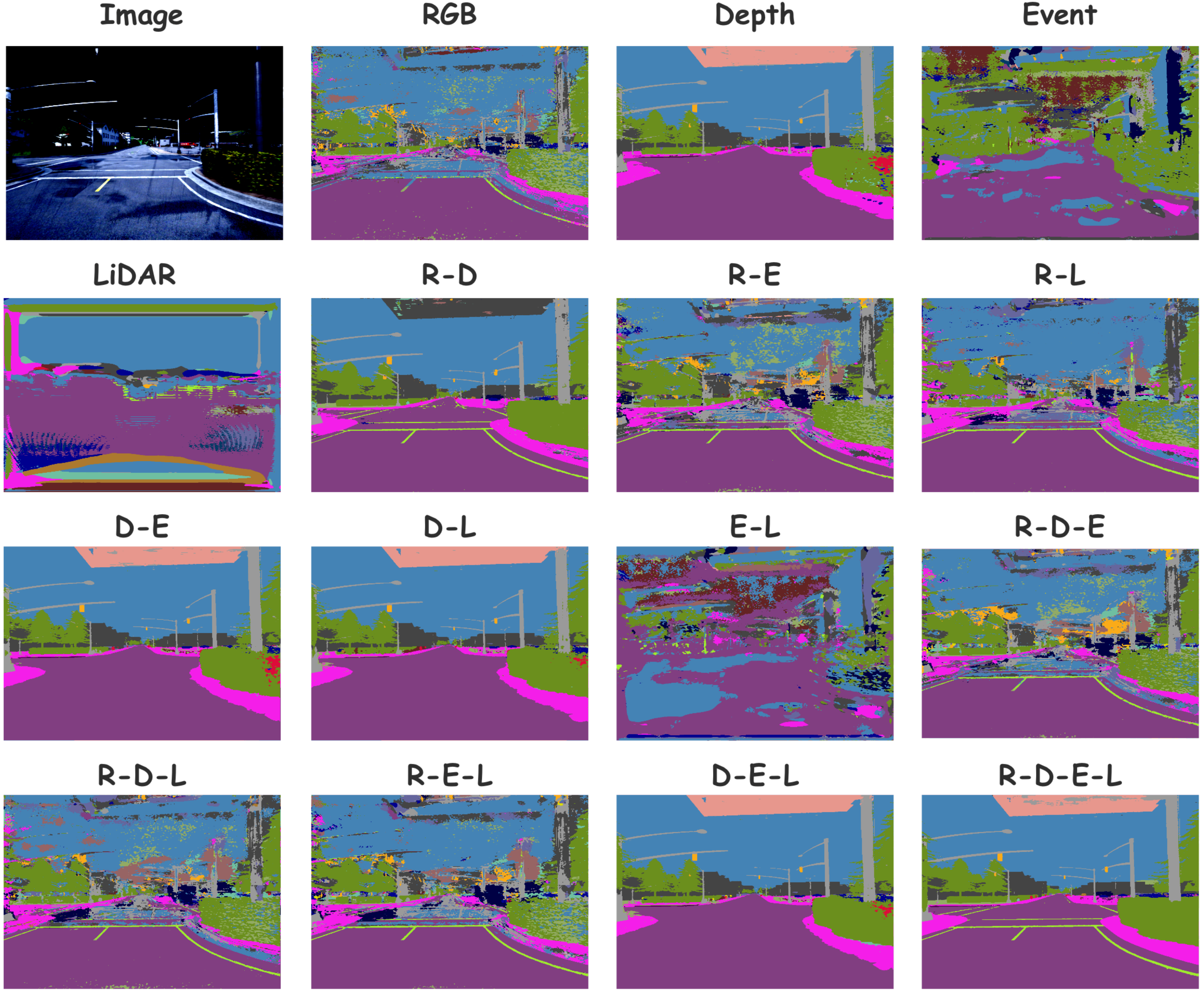}
  \caption{Supplementary Visualization of our EGFormer trained with four modalities under missing modality scenarios and
adverse weather conditions on DELIVER.}
  \label{val-full-deliver}
\end{figure}

\begin{figure}[t!]
  \centering
  \includegraphics[width=\linewidth]{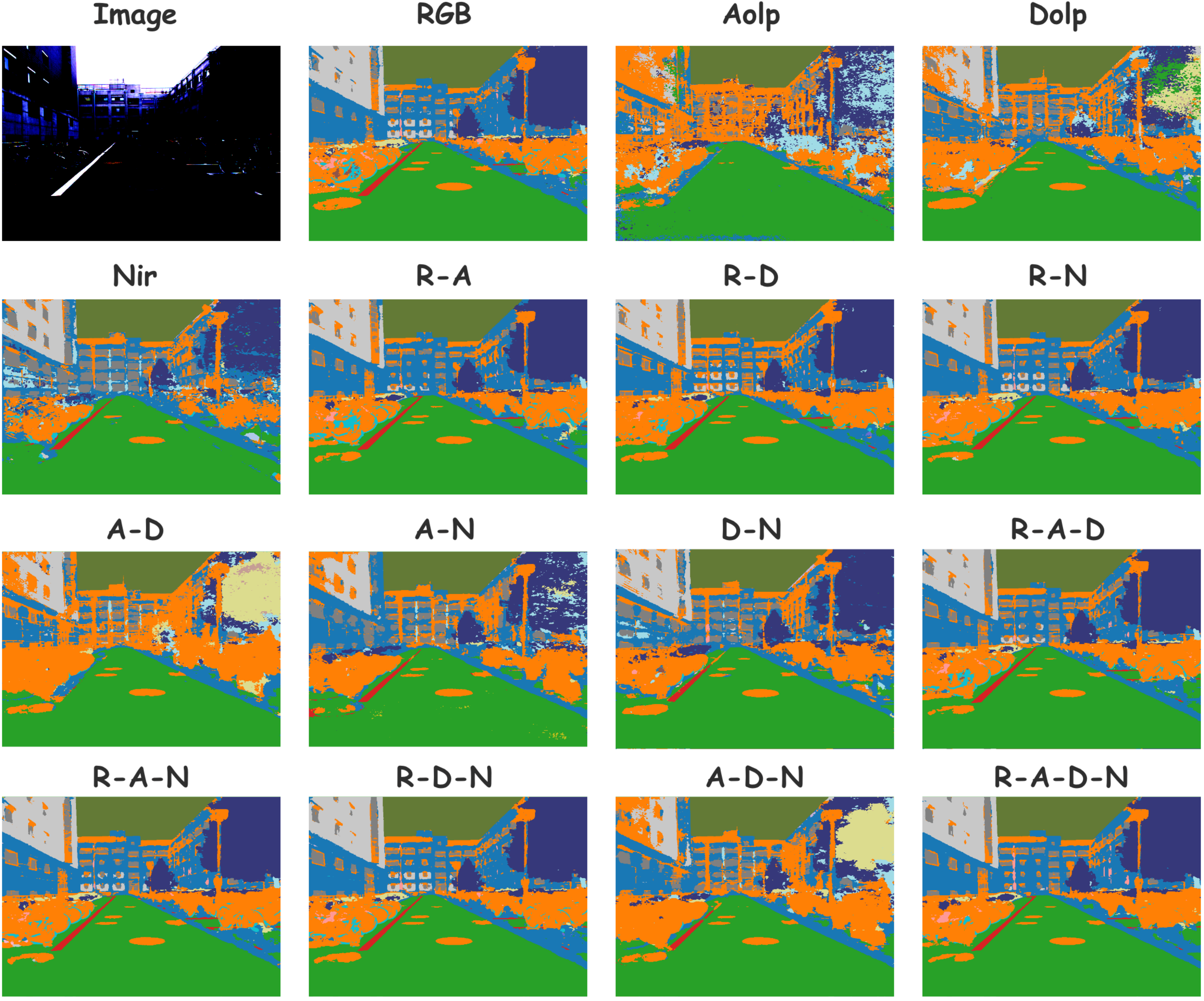}
  \caption{Supplementary Visualization of EGFormer trained with four modalities under missing modality scenarios and
adverse weather conditions on MCubeS.}
  \label{val-full-mcubes}
\end{figure}

Figure~\ref{val-full-deliver} showcases qualitative segmentation results on the DELIVER dataset using models trained with all four modalities (RGB, Depth, Event, LiDAR), but evaluated under various missing-modality scenarios. Each column corresponds to a specific combination of available inputs. Despite the absence of certain modalities, the predictions remain reasonably accurate and structurally consistent, especially when two or more modalities are present. Notably, combinations such as RD, RE, and RDE still preserve clear road boundaries and object contours under challenging lighting conditions, highlighting the strong generalization and robustness of EGFormer. Moreover, Figure~\ref{val-full-mcubes} presents similar visualizations on the MCubeS dataset, where the model is trained with all four modalities (RGB, AoLP, DoLP, NIR) and tested with different subsets. As shown, even with partial modality input, the network yields stable and precise segmentation outputs. For example, R-D-N and R-A-D-N preserve semantic boundaries and textures well, indicating the model’s ability to integrate complementary cues. These results demonstrate that EGFormer not only leverages full-modality information but also maintains reliable performance when some modalities are unavailable or degraded, making it suitable for real-world multimodal applications.

\begin{figure}
  \centering
  \includegraphics[width=\linewidth]{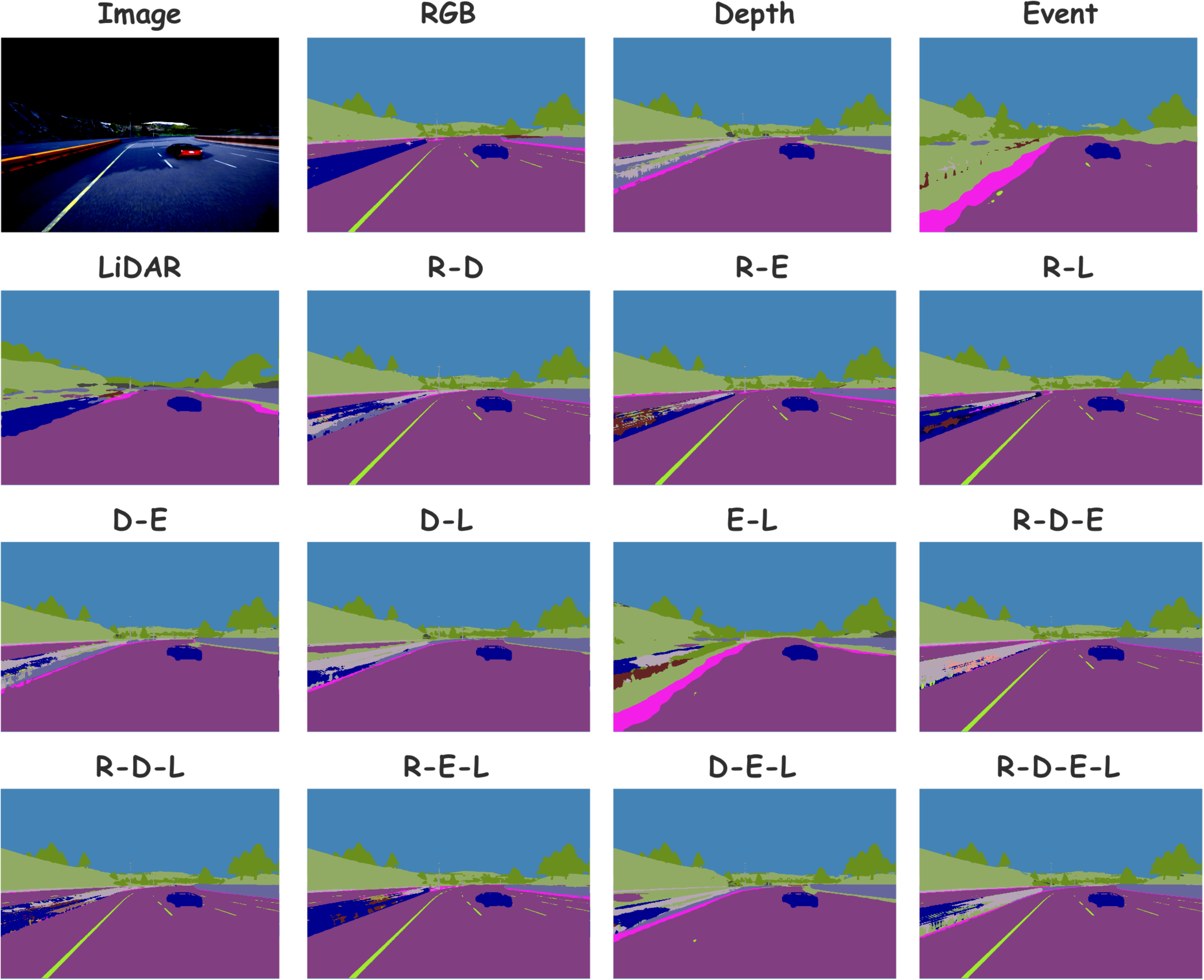}
  \caption{Supplementary visualization of EGFormer trained with all modalities on DELIVER.}
  \label{visual-full-deliver}
\end{figure}

\begin{figure}
  \centering
  \includegraphics[width=\linewidth]{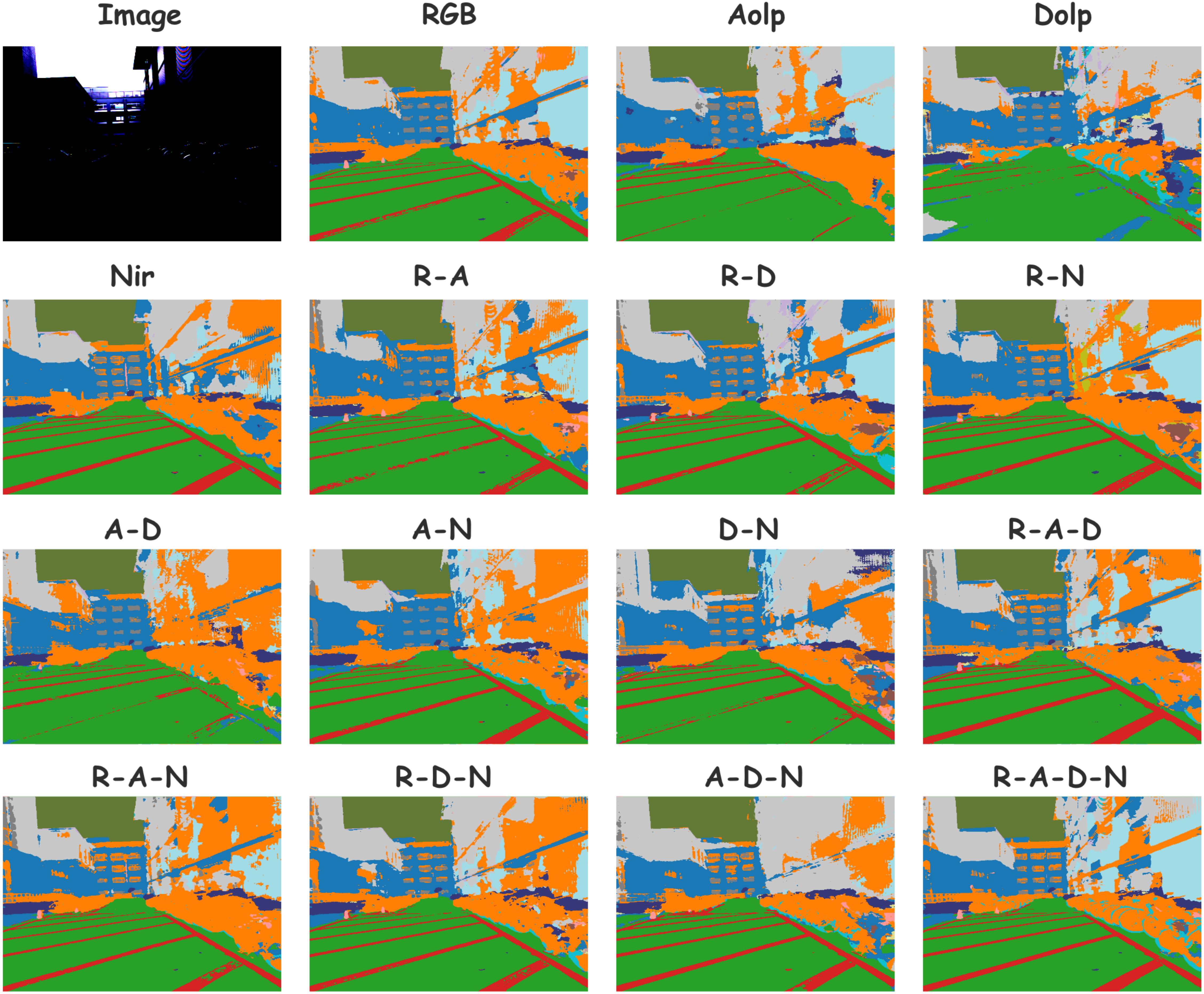}
  \caption{Supplementary visualization of EGFormer trained with all modalities on MCubeS.}
  \label{visual-full-mcubes}
\end{figure}

To further verify the robustness and generalizability of our EGFormer under realistic deployment scenarios, we evaluate the model trained with all available modalities on various modality combinations during inference. As illustrated in Figure~\ref{visual-full-deliver}, on the DELIVER dataset, the model exhibits strong segmentation performance even when some input modalities are missing. The predictions across different modality subsets (e.g., R-D, D-L, R-D-E) preserve semantic consistency and structural integrity, closely resembling the results from the full-modality input (R-D-E-L). This demonstrates that our method can effectively compensate for missing modalities via its learned cross-modal representations, making it suitable for challenging conditions such as sensor dropouts or adverse weather. Likewise, results on the MCubeS dataset (see Figure~\ref{visual-full-mcubes}) show that EGFormer trained with all four modalities (RGB, AoLP, DoLP, NIR) continues to generate high-quality segmentation outputs even under incomplete modality input, such as A-D-N or R-D-N. The stable visual performance across diverse input settings indicates that EGFormer not only benefits from multimodal fusion but also generalizes well in partial-modality scenarios. These findings reinforce the flexibility and reliability of EGFormer in real-world multimodal perception systems, where sensor availability can vary.


\end{document}